\documentclass{article}

    \usepackage[nonatbib, preprint]{tackling_climate_workshop_style}

\usepackage[utf8]{inputenc} 
\usepackage[T1]{fontenc}    
\usepackage{hyperref}       
\usepackage{url}            
\usepackage{booktabs}       
\usepackage{amsfonts}       
\usepackage{nicefrac}       
\usepackage{microtype}      
\usepackage[acronym]{glossaries}
\usepackage{newclude}
\usepackage{glossaries}
\usepackage{graphics}
\usepackage{graphicx}
\usepackage{wrapfig}
\usepackage{textcomp}
\usepackage[backend=biber, sorting=none]{biblatex}
\usepackage{caption, subcaption}
\newacronym{dmas}{DMAs}{District Metered Areas}
\newacronym{dhn}{DHA}{District Heating Network}
\newacronym{dhs}{DHS}{District Heating Systems}
\newacronym{wdn}{WDN}{Water Distrubution Network}
\newacronym{wdns}{WDNs}{Water Distrubution Networks}
\newacronym{cwt}{CWT}{Continous Wavelet Transform}
\newacronym{dl}{DL}{Deep Learning}
\newacronym{ml}{ML}{Machine Learning}
\newacronym{anns}{ANNs}{Artificial Neural Networks}
\newacronym{ann}{ANN}{Artificial Neural Network}
\newacronym{ai}{AI}{Artificial Intelligence}
\newacronym{rnn}{RNNs}{Recurrent Neural Networks}
\newacronym{sari}{SARIMAX}{Seasonal Auto-Regressive Integrated Moving Average with eXogenous factors}
\newacronym{cnn}{CNNs}{Convolutional Neural Networks}
\newacronym{DA}{da}{Day Ahead}
\newacronym{ID}{id}{Inter Day}
\newacronym{rmse}{RMSE}{Root Mean Squared Error}
\newacronym{mse}{MSE}{Root Mean Squared Error}
\newacronym{mae}{MAE}{Mean Absolute Error}
\newacronym{mape}{MAPE}{Mean Absolute Percentage Error}
\newacronym{smae}{SMAE}{Seasonal Mean Absolute Error}
\newacronym{mlp}{MLP}{Multi-Layer Perceptron}

\title{Advancing Heat Demand Forecasting with Attention Mechanisms: Opportunities and Challenges}

\author{%
Adithya Ramachandran$^{1,a}$ \thanks{Webpage: \url{https://lme.tf.fau.de/person/ramachandran/}}  , Thorkil Flensmark B. Neergaard$^{2,b}$,\\
  \textbf{Andreas Maier}$^{3,a}$, \textbf{Siming Bayer}$^{4,a}$\\
  $^a$Pattern Recognition Lab, Friedrich-Alexander-Universität Erlangen-Nürnberg\\
  Martensstr. 3, 91058 Erlangen, Germany\\
  $^b$Brønderslev Forsyning, Virksomhedsvej 20, 9700 Brønderslev, Denmark\\
  \texttt{$^1$adithya.ramachandran@fau.de,
  $^2$tbn@bronderslevforsyning.dk},\\
  \texttt{$^3$andreas.maier@fau.de, $^4$siming.bayer@fau.de}\\
}

\begin{document}

\maketitle
\begin{abstract}

Global leaders and policymakers are unified in their unequivocal commitment to decarbonization efforts in support of Net-Zero agreements.  
\acrfull{dhs}, while contributing to carbon emissions due to the continued reliance on fossil fuels for heat production, are embracing more sustainable practices albeit with some sense of vulnerability as it  could constrain their ability to adapt to dynamic demand and production scenarios.
As demographic demands grow and renewables become the central strategy in decarbonizing the heating sector, the need for accurate demand forecasting has intensified.
Advances in digitization have paved the way for \acrfull{ml} based solutions to become the industry standard for modeling complex time series patterns. In this paper, we focus on building a \acrfull{dl} model that uses deconstructed components of independent and dependent variables that affect heat demand as features to perform multi-step ahead forecasting of head demand. The model represents the input features in a time-frequency space and uses an attention mechanism to generate accurate forecasts.
The proposed method is evaluated on a real-world dataset and the forecasting performance is assessed against LSTM and CNN-based forecasting models. Across different supply zones, the attention-based models outperforms the baselines quantitatively and qualitatively, with an \acrfull{mae} of $0.105 \pm 0.06 kWh$ and a \acrfull{mape} of $5.4\% \pm 2.8 \%, $ in comparison the second best model with a \acrshort{mae} of $0.10 \pm 0.06 kWh$ and a \acrshort{mape} of $5.6\% \pm 3\%$.

\end{abstract}
\section{Introduction}

As the effects of climate change become increasingly pronounced, global policymakers are intensifying their efforts to combat these issues through climate-friendly policies \cite{european2012energy}. A major focus of these initiatives is the pursuit of carbon neutrality by decarbonizing energy systems within sustainable energy frameworks. Despite progress in renewable energy technologies, fossil fuels still provide 50\% of Europe’s primary energy, with 63\% of the residential sector's final energy consumption used for heating \cite{paardekooper2018heat}. District heating networks have significant potential for decarbonization through the integration of renewable energy sources \cite{steiner2015district}, underscoring the importance of precise heat demand forecasting methods. Such accuracy is crucial for efficient resource management, reducing energy waste, and facilitating the successful incorporation of renewables into existing energy systems.

With the growing proliferation of smart devices, time series modeling has greatly advanced, benefiting from the richer data streams and the progress in Machine Learning (ML) and Deep Learning (DL). While these advancements have led to sophisticated models, such as attention mechanisms that enable a model to prioritize and focus on the most critical parts of the input data, traditional statistical methods like ARIMA and SARIMAX remain popular \cite{DOTZAUER2002277}, \cite{FANG2016544}, \cite{chatterjee2021prediction}. These methods are valued for their intuitive additive and multiplicative approaches, offering a level of interpretability that complements the complexity of newer techniques. As attention-based models \cite{vaswani2023attentionneed}, including transformer models for time series, Large Language Models (LLMs) with embedded context, and foundation models for time series forecasting \cite{das2024decoderonlyfoundationmodeltimeseries}, become more prevalent, they still rely on well-crafted inputs to learn contextual representations effectively. Particularly with heat demand data, which exhibits strong daily and weekly seasonality along with yearly trends dictated by weather conditions, combining intuitive feature aggregation with these advanced models enhances their ability to capture the residual fluctuations that characterize time series data. This synergy between traditional and modern approaches is crucial for developing industrial applications aimed at reducing carbon emissions.

Given the widespread use of statistical modeling for its interpretability and the complex representations learned by DL models, we propose a DL-based forecasting approach that combines that employs an attention mechanism and incorporates individual components of decomposed time series as input features, allowing the model to capture the nuanced factors that influence heat consumption. By elucidating these individual components, we identify the key drivers of next-day heat demand. The effectiveness of the proposed approach is thoroughly evaluated using real-world heat demand data over an entire year, providing insights into its performance across different phases of the annual trend.

\section{Methodology}

\begin{figure}[htbp]
\centering
\includegraphics[width=1.0\linewidth]{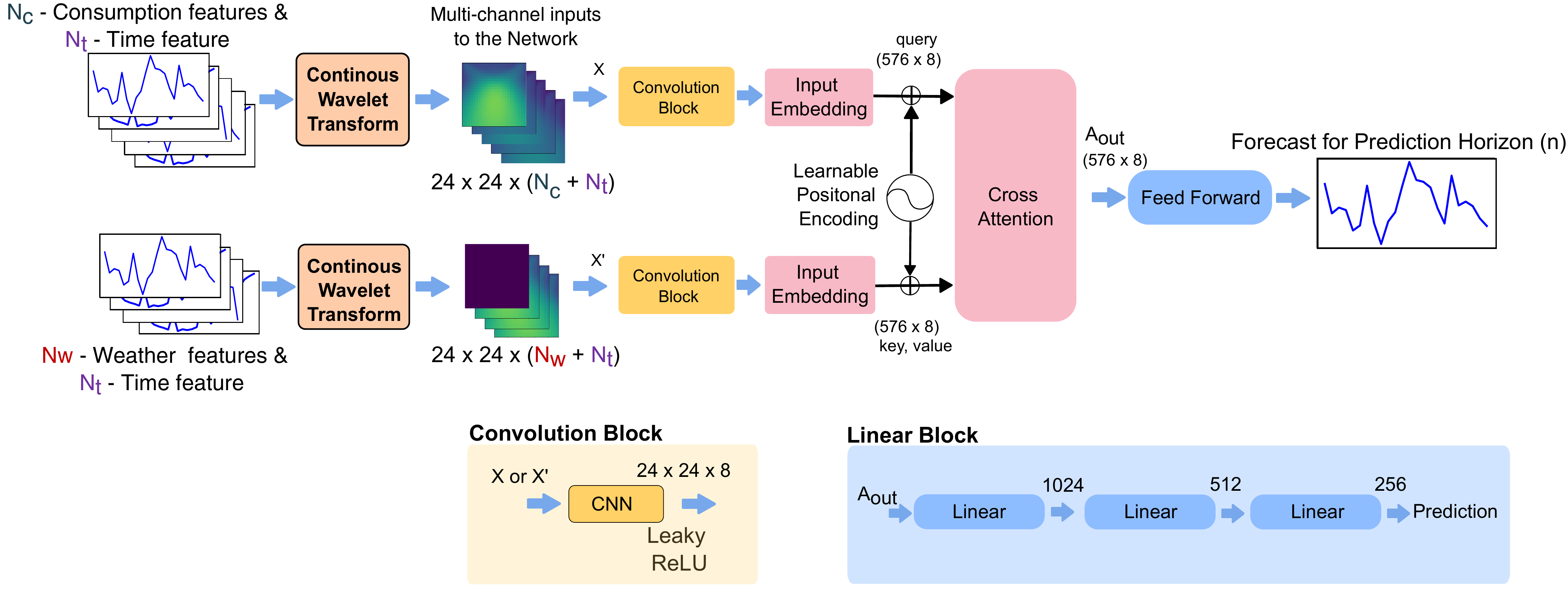}

\caption{Wavelet-based forecasting network with cross attention between the primary demand features and the supporting exogenous features. $N_{c}$, $N_{w}$, and $N_{t}$ represent the number of consumption, weather and time based features.}
\label{Fig1:block_diagram}
\end{figure}

For the downstream task of forecasting heat demand, we adopt and adapt a convolution-based network $F$   \cite{chatterjee2022heat} for which each input feature is represented in a Time-Frequency domain in the form of wavelet scalograms obtained through Continuous Wavelet Transform (CWT). During convolution, as different types of features such as demand and weather are concatenated to form a multi-channel input, the kernel learns the inter-dependency between each channel, however, there is a risk of features obscuring essential patterns. As a remedy, the architecture is modified to $F^{'}$ where the endogenous and exogenous features are embedded in two distinct branches, with a cross-attention block introduced after the convolutional layers, as illustrated in Figure \ref{Fig1:block_diagram}. However, this delays the model's ability to learn certain inter-dependencies across feature families early on, and the attention block compensates by dynamically focusing on and prioritizing the most relevant contexts from each branch during the merging process. With fewer and similar features, we only require a single convolution layer per branch to capture the essential patterns.

\begin{figure}[htbp]
\centering
\includegraphics[width=1.0\linewidth]{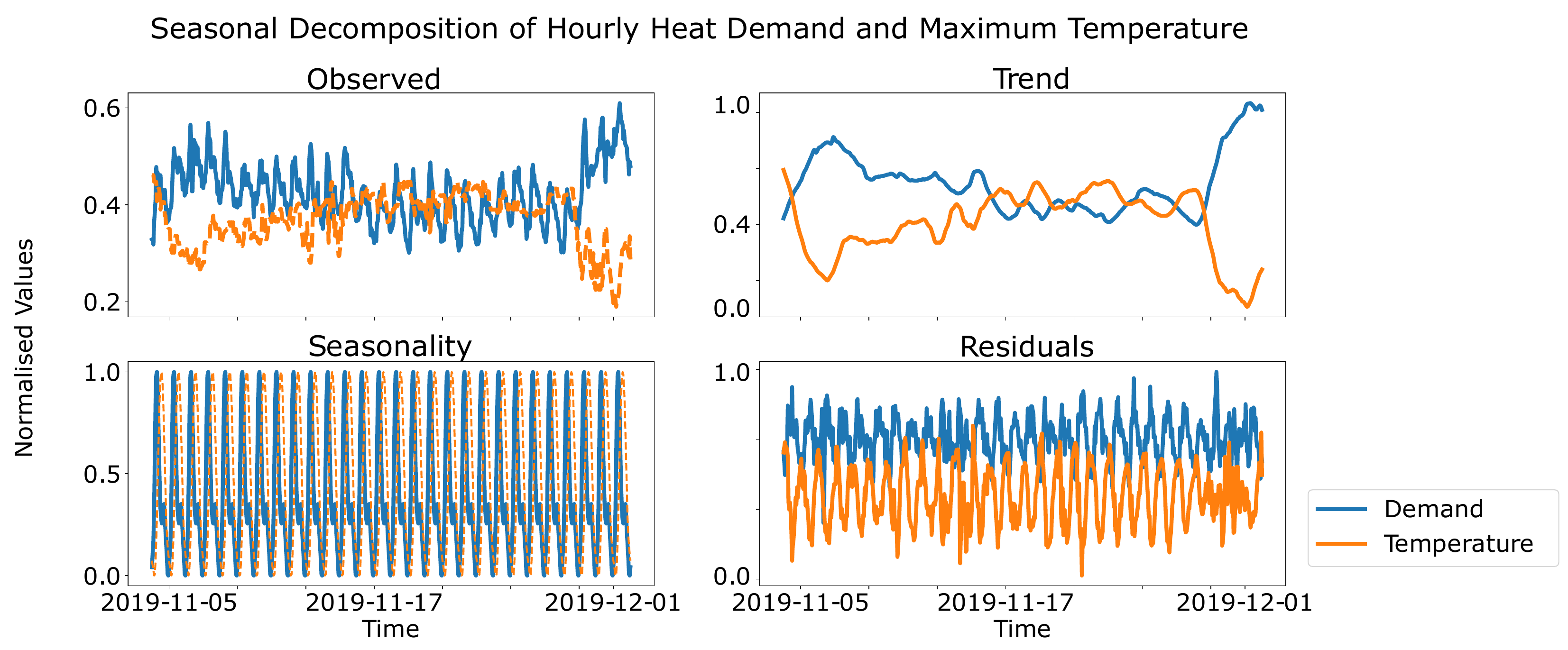}
\caption{Seasonal decomposition of demand data and maximum temperature sampled hourly for a specific DMA.}
\label{Fig2:SD}
\end{figure}

The endogenous features consist of historical consumption data, while the exogenous features include weather forecasts and time-based variables. Owing to the unidirectional influence of weather on demand patterns endogenous feature embedding is used as the query vector for the attention layer, while the exogenous features provide contextual support, enriching the representation of heat demand \cite{Shih2019},\cite{adiferra}. The attention layer is followed by a series of fully connected layers to generate the demand forecast. During feature selection, in addition to relevant features in their current form selected through correlational analysis, we also perform seasonal decomposition of the features, to obtain trend, seasonal, and residual components. Figure \ref{Fig2:SD} illustrates the decomposition of historical demand and observed maximum temperature. The temperature trend is selected as a feature due to its inverse influence on the trend of heat demand. However, the seasonal component of temperature is disregarded, while the seasonal component of demand is retained to reflect the DMA's daily or weekly patterns. Residual components are included as well, given there are identifiable patterns.

To forecast for heat demand \(\mathbf{y}(t) = [x_{t+1}, x_{t+2}, ..., x_{t+n}]\) at time $t$, where \(n=24\) represents the forecasting horizon, we discretize all temporal features with a winow of $h=24$ historical observations. These observations are then transformed using CWT to generate scalograms with dimensions of $ h \times s \times 1$, where $s=24$ represents the number of scales associated with the CWT. The scalograms are concatenated based on the family of the feature (demand or weather), forming the inputs to the model for predicting $y(t)$.

\section{Experimental Setup} \label{exp}

For the retrospective experiments, hourly heat consumption data from 2016 to 2020 for three DMAs from a Danish utility is used and actual observations in-place weather forecast. Apart from historical demand data at lags $24$, and $168$, maximum temperature, feels-like temperature, day of week encoded using sine and cosine function, along with their trend, seasonality and residuals components form the feature set. Experiments are conducted across three DL models. An LSTM model with four layers and 32 hidden units is the baseline for comparing the adopted wavelet scalogram-based model $F$ and modified model $F^{'}$. The entire year of 2019 is used for testing, while the rest follows a 80:20 training-validation split. Additional experimental settings are described in Appendix \ref{exps}.

\section{Results and Discussion}

\begin{figure}[htbp] 
    \centering
    \begin{subfigure}[t]{\linewidth} 
        \includegraphics[width=\linewidth]{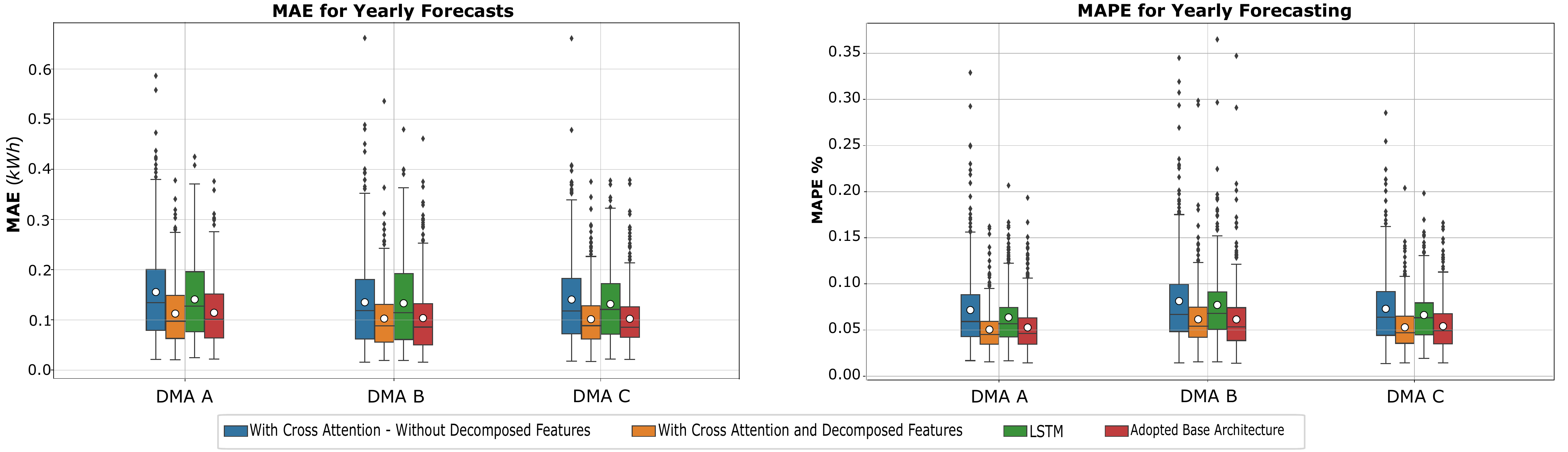} 
        \label{Fig:mape}
    \end{subfigure}

    \begin{subfigure}[t]{0.48\linewidth} 
        \includegraphics[width=\linewidth]{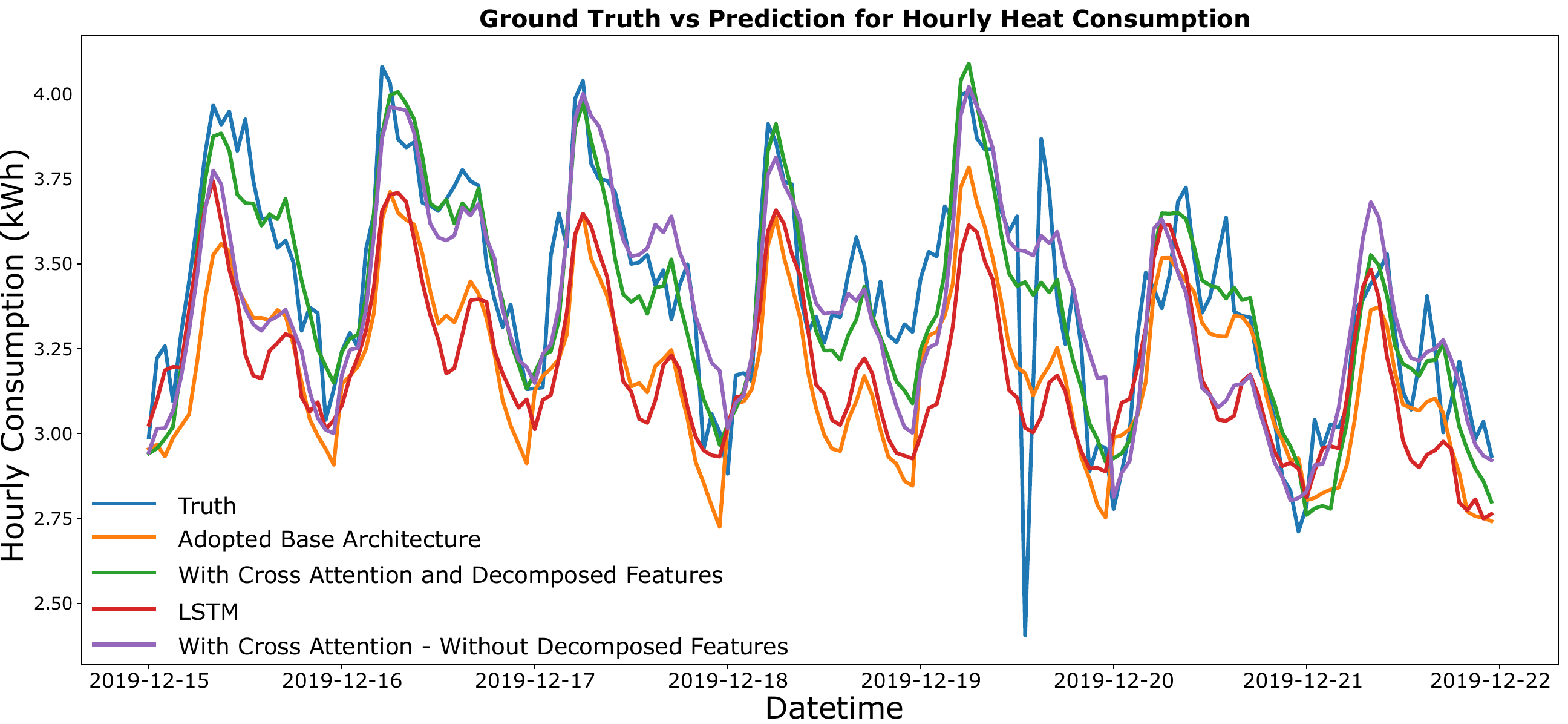} 
    \end{subfigure}
    \begin{subfigure}[t]{0.48\linewidth} 
        \includegraphics[width=\linewidth]{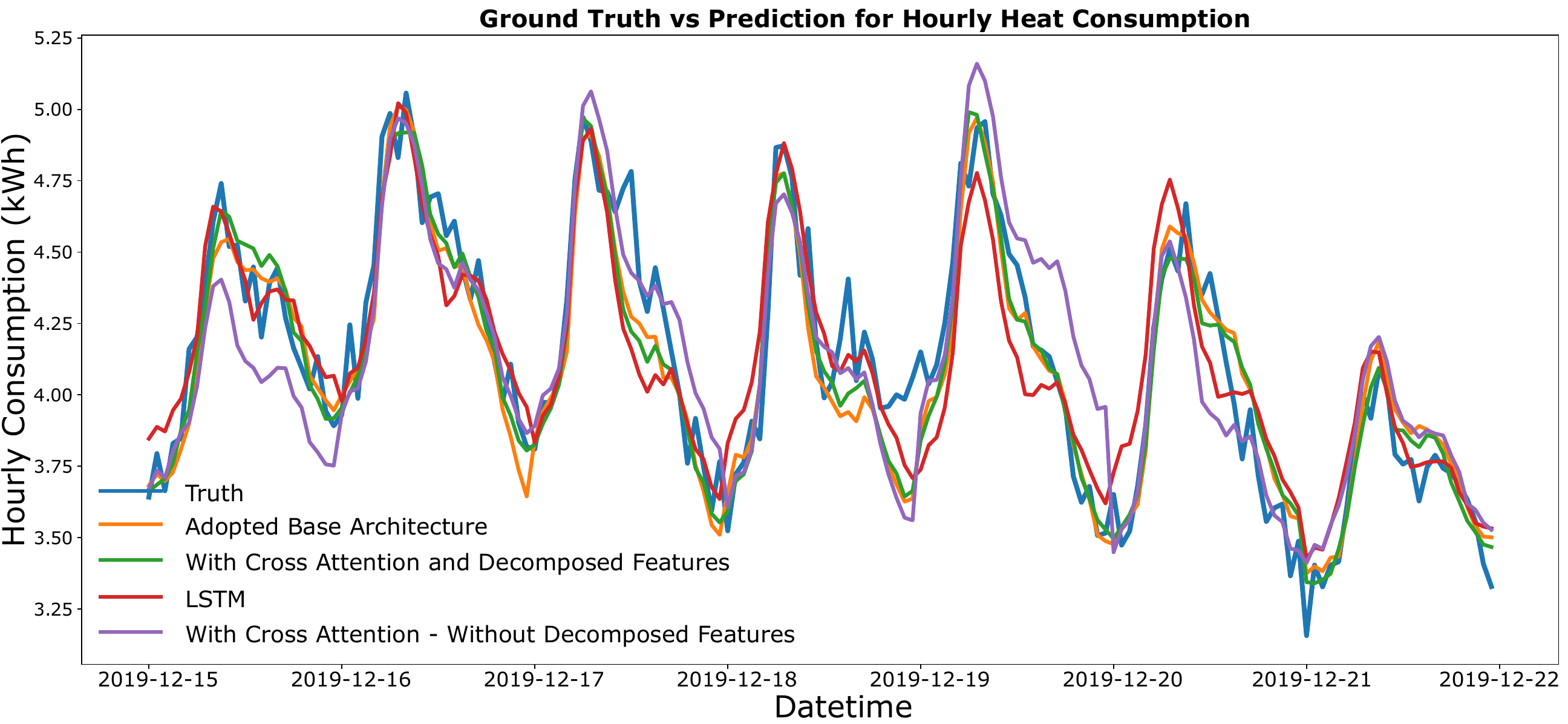} 

    \end{subfigure}
    \caption{Quantitative evaluation for the year 2019 - MAE (top left), and MAPE (top right) across DMAs; Qualitative plots of forecasts over a week in December 2019 for \textit{DMA A} and \textit{DMA B}.}

    \label{Fig:metrics_plots}
\end{figure}

The proposed framework is evaluated quantitatively and qualitatively against the baseline models and the results are depicted in Figure \ref{Fig:metrics_plots}. From the quantitative perspective, the method incorporating wavelet outperforms the LSTM model as expected. To quantify the impact of using decomposed components of the feature, the model $F^{'}$ was trained with and without the decomposed components as part of its feature set. It is evident that the model $F^{'}$ benefits with decomposed components as part of the input feature. Between $F$ and $F^{'}$, both models boosts superior forecasting abilities, $F^{'}$ consistently outperforms the other. $F^{'}$ also exhibits a lower degree of variance in comparison to the other models highlighting consistency with forecasting performance.

The qualitative results for \textit{DMA A} and \textit{B}, shows the ability of the model $F^{'}$  to follow daily and weekly patterns in demand closely than the other models. The baseline models  underforecasts the demand during the weekdays for \textit{DMA A}, while the proposed model captures the actual demand accurately throughout the week across both the DMAs. The model also show robustness to the perturbation in demand seen on the fifth day of \textit{DMA A}, by providing an accurate forecast for the sixth day. 


These results demonstrate the effectiveness of our proposed model in accurately and consistently forecasting heat demand by learning inter-and-intra feature dependencies weighted through a cross-attention block. Additionally, the proposed methodology reduces the number of trainable parameters of the $F$ by $97\%$ from approximately 155 million to approximately 5.7 million parameters, without losing forecasting performance. The model parameters are listed in Appendix \ref{mp}.

\section{Conclusion}

In this study, we introduced a robust framework for forecasting heat demand using a convolution-based model that integrates an attention mechanism and decomposed time series components. 
Our experiments showed that incorporating cross-attention between endogenous and exogenous features significantly improves the model's ability to learn feature-specific patterns and dynamically integrate contextual information. Quantitative evaluations confirmed that, especially with decomposed features, our model consistently outperforms traditional methods like LSTM and baseline wavelet-based models, achieving higher accuracy, reduced forecast variance, and a $97\%$ reduction in trainable parameters.
The architecture facilitates the integration of prior knowledge about demand-influencing factors into feature-specific branches and merges critical information through attention. Additionally, the forecast can be subjected to seasonal decomposition, allowing for intuitive comparisons between the forecasted and actual components, such as trends, seasonality, and residuals, to better understand the model's limitations and to address them with techniques such as regularisation. We believe that this approach, which emphasizes the identification and utilization of meaningful features, not only enhances model performance but also bridges the gap between research and practical application in the field of sustainable solutions.

\printbibliography 

@inproceedings{chatterjee2022heat,
  title={Heat Demand Forecasting with Multi-Resolutional Representation of Heterogeneous Temporal Ensemble},
  author={Chatterjee, Satyaki and Ramachandran, Adithya and Neergaard, Thorkil Flensmark and Maier, Andreas K and Bayer, Siming},
  booktitle={NeurIPS 2022 Workshop on Tackling Climate Change with Machine Learning},
  url={https://www.climatechange.ai/papers/neurips2022/46},
  year={2022}
}

@INPROCEEDINGS{cwt,
  author={Zhao, Yi and Shen, Yanyan and Zhu, Yanmin and Yao, Junjie},
  booktitle={2018 IEEE International Conference on Data Mining (ICDM)}, 
  title={Forecasting Wavelet Transformed Time Series with Attentive Neural Networks}, 
  year={2018},
  volume={},
  number={},
  pages={1452-1457},
  doi={10.1109/ICDM.2018.00201}}

@misc{vaswani2023attentionneed,
      title={Attention Is All You Need}, 
      author={Ashish Vaswani and Noam Shazeer and Niki Parmar and Jakob Uszkoreit and Llion Jones and Aidan N. Gomez and Lukasz Kaiser and Illia Polosukhin},
      year={2023},
      eprint={1706.03762},
      archivePrefix={arXiv},
      primaryClass={cs.CL},
      url={https://arxiv.org/abs/1706.03762}, 
}

@article{paardekooper2018heat,
  title={Heat Roadmap United Kingdom: Quantifying the Impact of Low-Carbon Heating and Cooling Roadmaps},
  author={Paardekooper, Susana and Lund, Rasmus S{\o}gaard and Mathiesen, Brian Vad and Chang, Miguel and Petersen, Uni Reinert and Grundahl, Lars and David, Andrei and Dahlb{\ae}k, Jonas and Kapetanakis, Ioannis Aristeidis and Lund, Henrik and others},
  year={2018}
}

@article{steiner2015district,
  title={District Energy in Cities: Unlocking the Potential of Energy Efficiency and Renewable Energy},
  author={Steiner, A and Yumkella, KK and Clos, J and Begin, GV},
  journal={United Nations Environment Programme (UNEP): Nairobi, Kenya},
  year={2015}
}

@book{european2012energy,
  title={Energy roadmap 2050},
  author={European Commission},
  year={2012},
  publisher={Publications Office of the European Union}
}

@article{DOTZAUER2002277,
title = {Simple model for prediction of loads in district-heating systems},
journal = {Applied Energy},
volume = {73},
number = {3},
pages = {277-284},
year = {2002},
issn = {0306-2619},
doi = {https://doi.org/10.1016/S0306-2619(02)00078-8},
url = {https://www.sciencedirect.com/science/article/pii/S0306261902000788},
author = {Erik Dotzauer}
}

@article{FANG2016544,
title = {Evaluation of a multiple linear regression model and SARIMA model in forecasting heat demand for district heating system},
journal = {Applied Energy},
volume = {179},
pages = {544-552},
year = {2016},
issn = {0306-2619},
doi = {https://doi.org/10.1016/j.apenergy.2016.06.133},
url = {https://www.sciencedirect.com/science/article/pii/S0306261916309217},
author = {Tingting Fang and Risto Lahdelma}
}

@inproceedings{chatterjee2021prediction,
  title={Prediction of Household-level Heat-Consumption using PSO enhanced SVR Model},
  author={Chatterjee, Satyaki and Bayer, Siming and Maier, Andreas K},
  booktitle={NeurIPS 2021 Workshop on Tackling Climate Change with Machine Learning},
  url={https://www.climatechange.ai/papers/neurips2021/42},
  year={2021}
}

@Article{Shih2019,
author={Shih, Shun-Yao
and Sun, Fan-Keng
and Lee, Hung-yi},
title={Temporal pattern attention for multivariate time series forecasting},
journal={Machine Learning},
year={2019},
month={9},
day={01},
volume={108},
number={8},
pages={1421-1441},
issn={1573-0565},
doi={10.1007/s10994-019-05815-0},
url={https://doi.org/10.1007/s10994-019-05815-0}
}

@inproceedings{adiferra,
 author = {Ramachandran, Adithya and Mousa, Hatem and Maier, Andreas and Bayer, Siming},
 booktitle = {WDSA CCWI 2024 - 3rd International Joint Conference on Water Distribution Systems Analysis and Computing and Control for the Water Industry},
 date = {2024-07-01/2024-07-05},
 faupublication = {yes},
 peerreviewed = {unknown},
 title = {{A} {Week} {Ahead} {Water} {Demand} {Forecasting} using {Convolutional} {Neural} {Network} on {Multi}-{Channel} {Wavelet} {Scalogram}},
 venue = {University of Ferrara, Ferrara, Italy},
 year = {2024}
}

@misc{das2024decoderonlyfoundationmodeltimeseries,
      title={A decoder-only foundation model for time-series forecasting}, 
      author={Abhimanyu Das and Weihao Kong and Rajat Sen and Yichen Zhou},
      year={2024},
      eprint={2310.10688},
      archivePrefix={arXiv},
      primaryClass={cs.CL},
      url={https://arxiv.org/abs/2310.10688}, 
}

\newpage
\section{Appendix}

\subsection{Base Architecture}
\begin{figure}[htbp]
\centering
\includegraphics[width=1.0\linewidth]{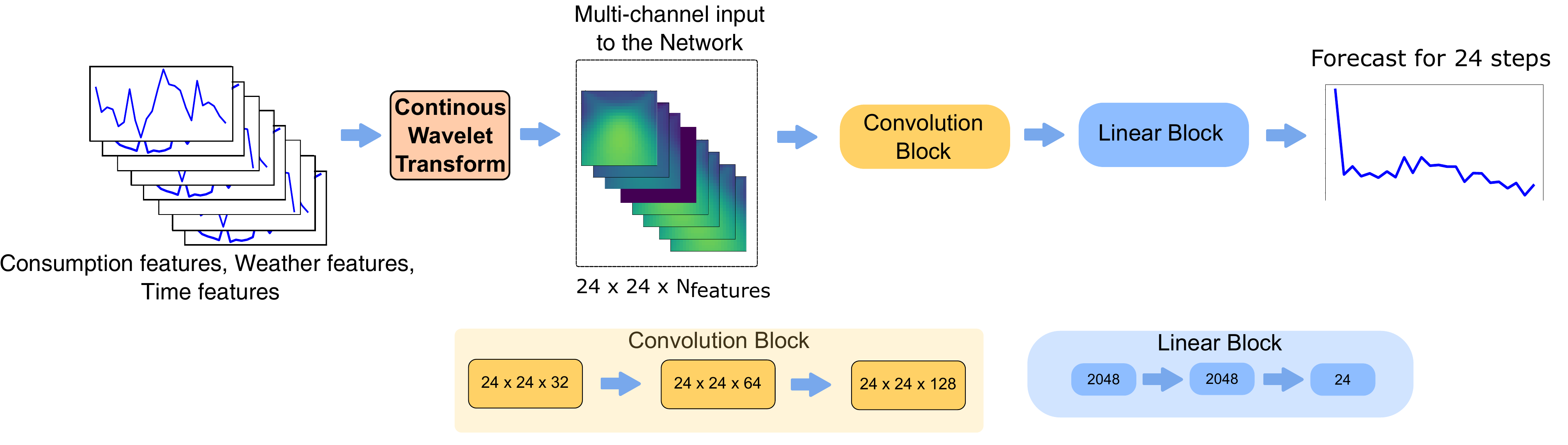}
\caption{Wavelet scalogram-based architecture used for heat demand forecast as adopted from \cite{chatterjee2022heat}.}
\label{Fig6_1:originialarch}
\end{figure}

\subsection{Experimental Parameters} \label{exps}
Extension to the experimental setup discussed in Section \ref{exp}.
\subsubsection{Data Preprocessing}
The demand data for various DMAs is obtained by aggregating individual smart meter readings within each DMA and is differenced to achieve stationarity. The data undergoes preprocessing to remove statistical outliers, as well as to address missing values or instances of negative consumption. The dataset is segmented into $h = 24$ hour intervals, spanning from midnight to midnight. Relevant features are identified through correlational analysis and seasonal decomposition. Due to the observed daily and weekly consumption patterns, two lagged consumption features—24 hours (previous day) and 168 hours (same day in the previous week)—are used as demand features. For weather features, the hourly maximum temperature and feels-like temperature for the forecast day are selected. Time-based features, such as the hour of the day and day of the week, are cyclically encoded using sine and cosine functions.

\subsubsection{Training parameters}
All models are trained using the ADAM optimizer with \acrshort{mse} as the loss function to convergence, with an early stopping criterion to prevent overfitting. Hyperparameters are further fine-tuned using grid search. A batch size of $256$ and a learning rate of $0.01$ are applied consistently across all models during training. For the model $F^{'}$ visualised in Figure \ref{Fig1:block_diagram} the positional encoding of tokens is managed by a learnable layer, chosen based on prior experiments comparing it with sinusoidal encoding.

\subsection{Model Parameters} \label{mp}
The number of trainable parameters in the LSTM model, the baseline wavelet model $F$, and the proposed model $F^{'}$ is given below:

\begin{itemize}
    \item LSTM: 32,536
    \item  Scalogram - Base model $F$: 155,339,512 
    \item  Wavelet Scalogram - with Cross Attention $F^{'}$:  5,789,152 

\end{itemize}

\end{document}